\documentclass[runningheads]{llncs}
\usepackage{subcaption}
\usepackage{pgfplots}
\usepackage[table]{xcolor}
\pgfplotsset{compat=1.17}
\usepackage{booktabs} 
\usepackage{amsmath}
\usepackage{amsfonts}
\usepackage[T1]{fontenc}
\usepackage{graphicx}
\begin{document}
\title{Feature-level Interaction Explanations in Multimodal Transformers}
\author{
Yeji Kim\inst{1} \and
Housam Babiker\inst{2} \and
Mi-Young Kim\inst{3} \and
Randy Goebel\inst{1}
}

\authorrunning{Y. Kim et al.}

\institute{
University of Alberta, Edmonton, AB, Canada \\
\email{\{yeji7,rgoebel\}@ualberta.ca}
\and
University of Waikato, Hamilton, New Zealand \\
\email{housam.babiker@waikato.ac.nz}
\and
University of Alberta, Camrose, AB, Canada \\
\email{\{miyoung2\}@ualberta.ca}
}
\maketitle              % typeset the header of the contribution
\begin{abstract}
Multimodal Transformers often produce predictions without clarifying how different modalities jointly support a decision. Most existing multimodal explainable AI (MXAI) methods extend unimodal saliency to multimodal backbones, highlighting important tokens or patches within each modality, but they rarely pinpoint which cross-modal feature pairs provide complementary evidence (synergy) or serve as reliable backups (redundancy). We present Feature-level I$^2$MoE (FL-I$^2$MoE), a structured Mixture-of-Experts layer that operates directly on token/patch sequences from frozen pretrained encoders and explicitly separates unique, synergistic, and redundant evidence at the feature level. We further develop an expert-wise explanation pipeline that combines attribution with top-$K\%$ masking to assess faithfulness, and we introduce Monte Carlo interaction probes to quantify pairwise behavior: the Shapley Interaction Index (SII) to score synergistic pairs and a redundancy-gap score to capture substitutable (redundant) pairs. Across three benchmarks (MM-IMDb, ENRICO, and MMHS150K), FL-I$^2$MoE yields more interaction-specific and concentrated importance patterns than a dense Transformer with the same encoders. Finally, pair-level masking shows that removing pairs ranked by SII or redundancy-gap degrades performance more than masking randomly chosen pairs under the same budget, supporting that the identified interactions are causally relevant. Code is available at \url{https://github.com/dut0817/FL-I2MoE}.

\end{abstract}
\section{Introduction}

Multimodal models that combine diverse modalities (e.g., text and images) are increasingly deployed in high-stakes domains such as clinical decision support, legal analytics, and safety-critical content moderation~\cite{baltruvsaitis2018multimodal,sun2024review}. In these settings, a prediction is rarely trusted based on plausible pixels or words in isolation. What matters is how modalities jointly support the decision, either by providing complementary evidence or by offering redundant, mutually confirming signals that make predictions robust to noise or partial modality failure. However, conventional saliency maps can still look convincing even when the model effectively relies on a single modality, which obscures whether the decision is truly cross-validated or fragile~\cite{hessel2020does,parcalabescu2023mm,chaudhuri2025closer}.

\begin{figure}[t]
    \centering
    \includegraphics[width=\linewidth]{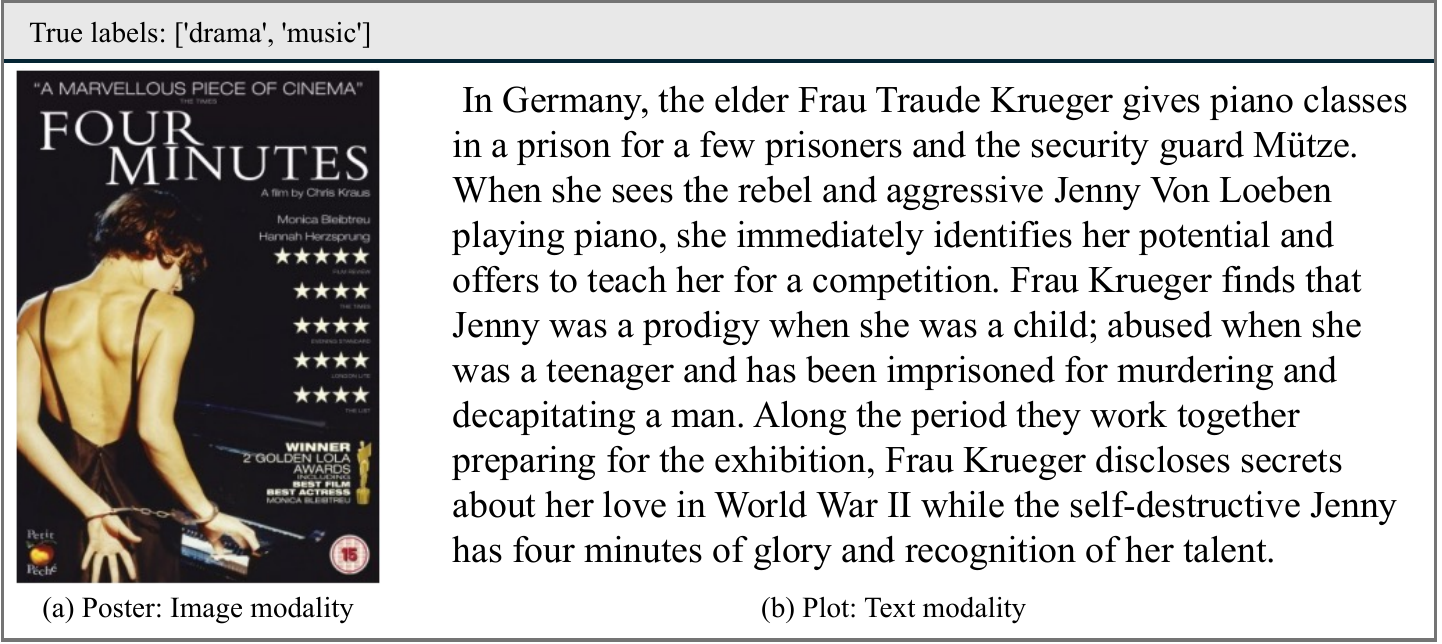}
    \caption{Input modalities for the qualitative MM-IMDb example:
    (a) movie poster image and (b) plot summary (\emph{Four Minutes},
    true labels: \textit{drama}, \textit{music}).}
    \label{fig:mmimdb_qual_input}
\end{figure}

Figure~\ref{fig:mmimdb_qual_input} illustrates this with a simple text-image example. Even for the same movie poster and plot, a model that predicts the genre by jointly using piano-related regions in the image and words such as ``piano'' or ``concert'' is behaving very differently from a model that mostly trusts the text, or one that guesses ``drama'' from a face alone. For practitioners, it is crucial to know whether each patch or token acts as unique evidence, whether a given patch-token pair serves as redundant backup, or whether the two together provide genuinely synergistic information. This concrete, feature interaction determines whether a prediction is robust because modalities cross-check each other, or fragile because it hinges on a single unreliable view.

Current multimodal explainable AI (MXAI) methods only partially address this need~\cite{jin2021one}. Most work extends unimodal attribution techniques to multimodal Transformers by computing importance scores over joint feature sequences or cross-attention layers and visualizing them as saliency maps. These explanations are useful for showing ``where the model looks,'' but they usually treat the backbone as a single network. They do not reveal what kind of interaction the highlighted features have or how they jointly support the prediction~\cite{jin2021one,sun2024review}. As a result, users cannot easily tell whether a prediction is well supported by consistent multimodal evidence or dominated by one potentially noisy source.

Game-theoretic and information-theoretic approaches aim to quantify multimodal contributions more explicitly. Shapley-value methods, including the Shapley Interaction Index (SII), can in principle model pairwise interactions, and PID-style decompositions separate information into unique, redundant, and synergistic components~\cite{grabisch1999axiomatic,wang2025multishap,parcalabescu2023mm,wollstadt2023rigorous}. Recent work like MultiSHAP applies Shapley-style interaction attribution post-hoc to a fixed multimodal backbone, but it typically operates at the modality level or over coarse feature groups and therefore does not expose fine-grained feature-level pairs~\cite{wang2025multishap}. Moreover, estimating Shapley-style pair interactions over dense token–patch spaces  becomes computationally prohibitive, which often necessitates aggregation or restricted candidate definitions.

Interaction-aware architectures partially address this gap by introducing explicit components for different interaction roles. In particular, I$^2$MoE~\cite{xin2025i2moe} incorporates PID-inspired experts (uniqueness, synergy, and redundancy) and an interaction loss to encourage specialization, but it operates on pooled modality vectors and therefore cannot reveal which feature-level cross-modal pairs each expert actually relies on. In this work, we revisit interaction-aware multimodal modeling from a feature-level explainability perspective: we extend I$^2$MoE to operate directly on token and patch sequences, and we use Monte Carlo interaction probes together with pair-level masking to quantitatively identify the cross-modal feature pairs that serve as synergistic or redundant evidence.

Our contributions are summarized as follows:
\begin{itemize}
    \item
    We introduce a Feature-level I$^2$MoE (FL-I$^2$MoE) that routes features through uniqueness, synergy, and redundancy experts, providing structured interaction-aware explanations while improving performance.

    \item
    We develop an expert-wise pipeline combining attribution with top-$K\%$ masking, and show improved faithfulness and interaction specificity over a dense Transformer on three benchmarks.
    
    \item
    We define cross-modal interaction scores (SII-based synergy and a redundancy-gap metric) and use pair-level masking to show that top-ranked feature pairs are causally important and aligned with expert roles, revealing concrete synergistic and redundant pair interactions.
    
\end{itemize}

\section{Related Work}

\subsection{Multimodal XAI and Feature Attribution}
Multimodal explainable AI (MXAI) seeks to make model predictions over text, images, speech, and tables more transparent~\cite{sun2024review}. In practice, most methods extend unimodal feature attribution to multimodal Transformers by computing attention- or gradient-based scores over joint feature sequences or cross-attention layers and visualizing them as saliency maps~\cite{abnar2020quantifying,sundararajan2017axiomatic,chefer2021generic}. While these approaches are widely used and often yield plausible explanations, they typically treat the backbone as a single monolithic network and do not explicitly distinguish unique, redundant, or synergistic cross-modal information~\cite{jin2021one}. As a result, they mainly reveal \emph{where} the model attends, but not \emph{how} concrete cross-modal feature pairs support or substitute for each other in the prediction.

\subsection{Quantifying Multimodal Contributions and Interactions}
Beyond saliency maps, game-theoretic and information-theoretic approaches aim to quantify multimodal contributions more explicitly. Shapley-value methods, including the Shapley Interaction Index (SII), can in principle model feature contributions and pairwise interaction effects~\cite{grabisch1999axiomatic,wang2025multishap,parcalabescu2023mm}, while Partial Information Decomposition (PID) separates information into unique, redundant, and synergistic components~\cite{wollstadt2023rigorous}. These tools provide notions of interaction and are often applied \emph{post-hoc} to an already trained backbone, including sample-level analyses such as MultiSHAP that attribute cross-modal effects for individual inputs~\cite{wang2025multishap}. However, existing work typically defines features at the modality level or over coarse feature groups, and the resulting interaction scores are frequently aggregated into global statistics. Moreover, estimating Shapley-style pair interactions over dense token-patch spaces quickly becomes computationally prohibitive, which often necessitates aggregation or restricted candidate definitions rather than exposing fine-grained feature-level pairs. Related perturbation-based analyses include EMAP~\cite{hessel2020does} and DIME~\cite{lyu2022dime}, which remain post-hoc to a fixed backbone.

\subsection{Interaction-aware Multimodal Architectures}

A separate line of work tackles multimodal interactions at the architectural level rather than only through post-hoc explanations. Classical multimodal models use modality-specific encoders followed by a single fusion module, which can entangle diverse interaction patterns in one shared representation. More recent architectures seek to structurally separate interaction types, for example via shared/private factorization~\cite{tsai2018learning} or dedicated modality-specific vs.\ cross-modal submodules. Interaction-aware architectures partially address the above limitations by introducing explicit components for different interaction roles, enabling a structural link between where interactions are represented and what type of cross-modal evidence they capture.

Interpretable Multimodal Interaction-aware Mixture-of-Experts (I$^2$MoE)~\cite{xin2025i2moe} is an end-to-end interaction-aware architecture that integrates PID-inspired experts (uniqueness, synergy, and redundancy) together with an interaction loss that encourages role specialization, and summarizes behavior through expert weights at the instance and dataset level. However, I$^2$MoE operates on pooled modality vectors, and therefore cannot expose which fine-grained feature pairs constitute synergistic or redundant evidence.

We build on I$^2$MoE by extending it to operate directly on token and patch sequences, and by coupling its interaction experts with quantitative interaction analyses. Using Monte Carlo interaction probes (SII-based synergy and a redundancy-gap score) together with pair-level masking, we identify and validate concrete cross-modal feature pairs that act as synergistic or redundant evidence.

\section{Method}\label{sec:method}
\subsection{Architecture and Training}
\subsubsection{Overview of I$^2$MoE}
I$^2$MoE is a multimodal Mixture-of-Experts architecture inspired by PID-style notions of unique, synergistic, and redundant information~\cite{xin2025i2moe}. Conceptually, unique information refers to predictive signal specific to a single modality and is not recoverable from the others. Redundant information corresponds to shared, mutually confirming evidence that appears in multiple modalities, so that either modality can support the decision on its own. In contrast, synergistic information captures complementary evidence that becomes predictive primarily when modalities are combined, meaning that the joint configuration provides more support than the sum of individual contributions.

I$^2$MoE instantiates these interaction roles via dedicated experts (uniqueness, synergy, and redundancy) together with a reweighting network $W$ that outputs non-negative mixture weights. Given encoded modality representations $\{e_m\}$, each expert produces a task prediction $\hat{y}_i$, and the final prediction is computed as a weighted mixture $\hat{y}=\sum_i w_i\hat{y}_i$. During training, the interaction loss $\mathcal{L}_{\text{int}}$ defined in~\cite{xin2025i2moe} is combined with the main task loss (cross-entropy) to encourage the experts to specialize toward their intended roles, allowing the model to route an input toward modality-specific evidence (unique experts), complementary cross-modal evidence (synergy expert), or shared substitutable evidence (redundancy expert), rather than entangling all behaviors in a single fusion module. We follow the original I$^2$MoE work and use exactly the same training objective $\mathcal{L}$, keeping the definition and hyperparameters of $\mathcal{L}_{\text{int}}$ unchanged, and treat our contribution as an architectural and analysis extension on top of this objective.

\subsubsection{Feature-level Encoders and Fusion Module}
\begin{figure}[t]
    \centering
    \includegraphics[width=\linewidth]{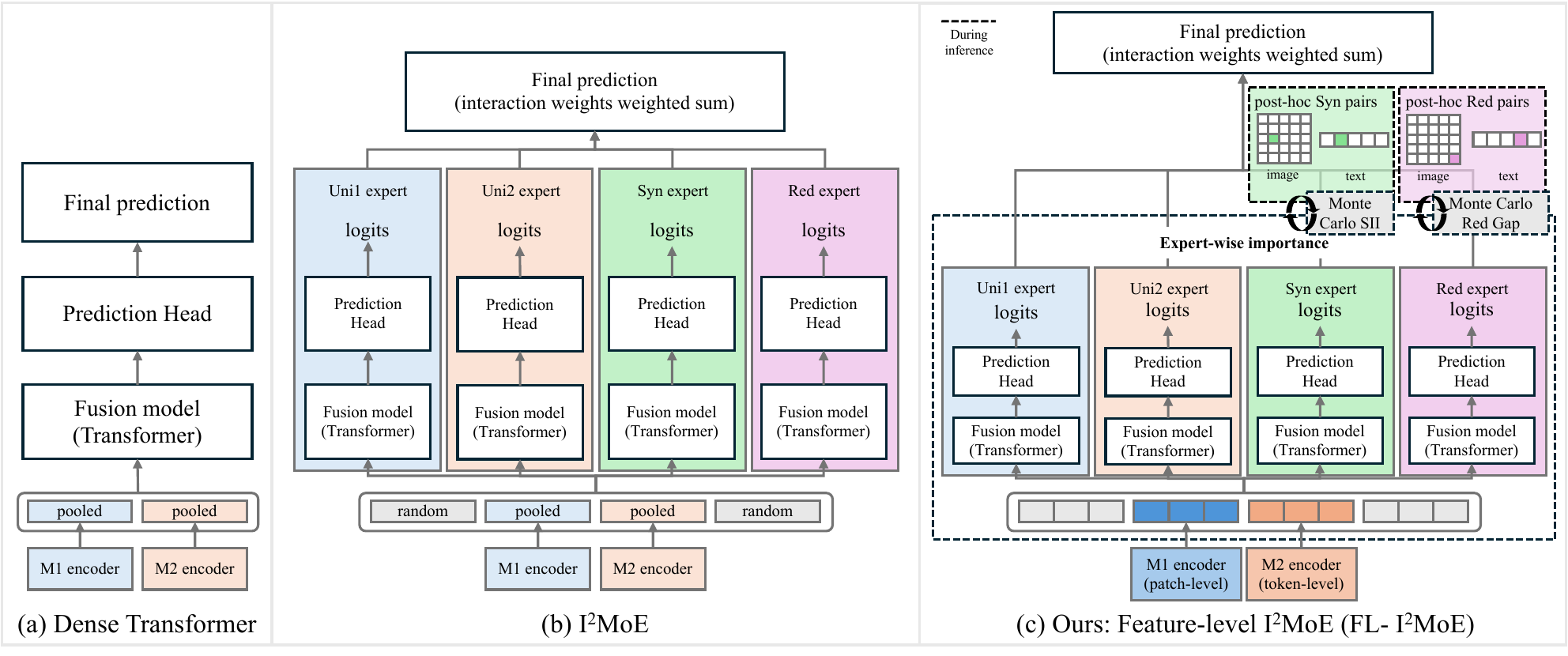}
    \caption{Overview of the compared architectures and our analysis pipeline.
    (a) dense Transformer with pooled modality encoders and
    a single fusion module.
    (b) Original I$^2$MoE with uniqueness, synergy, and redundancy experts
    operating on pooled modality vectors.
    (c) Proposed FL-I$^2$MoE: patch- and token-level encoders
    feed interaction experts, which output class logits for the main task. In the dashed region, at inference time, we derive expert-wise feature importance maps, propagate them to select cross-modal token-patch pairs, and then compute synergy and redundancy-gap scores to identify synergistic or redundant pairs.}
    \label{fig:overview}
\end{figure}

Figures~\ref{fig:overview}(a) and (b) summarize the dense Transformer
and the original I$^2$MoE, respectively.
Panel (c) shows our feature-level variant, where we replace pooled
encoders with patch- and token-level encoders while keeping the same
interaction experts and gating mechanism as in the original I$^2$MoE.
Throughout Figure~\ref{fig:overview}, we illustrate the architecture with two modalities for clarity, but the formulation and implementation naturally extend to an arbitrary number of modalities; in our experiments, we instantiate the model with two modalities on each dataset.

In the original I$^2$MoE, each modality input $x_m$ is encoded into a latent representation $e_m = E_m(x_m)$ and then pooled to produce a single vector per modality.
The interaction module receives the set of pooled vectors $\{e_m\}$ and produces expert-wise predictions.
In our feature-level variant, we replace this pooling step with sequence-preserving encoders that output feature-level embeddings
\[
E_m(x_m) = \{ h_{m,1}, \dots, h_{m,T_m} \}, \quad h_{m,t} \in \mathbb{R}^d,
\]
and we concatenate the modality-specific sequences into a joint sequence
\[
H = \big[ H^{(1)} \,\Vert\, H^{(2)} \,\Vert\, \dots \,\Vert\, H^{(M)} \big],
\]
where $H^{(m)} \in \mathbb{R}^{T_m \times d}$ and the sequence lengths $T_m$ may differ across modalities.
Concatenation is performed only along the sequence dimension; to ensure the fusion module is well-defined, we enforce a shared hidden size $d$ across modalities by mapping each encoder's native output to a common dimension via an output projection.
This yields a joint input of shape $(\sum_m T_m)\times d$ for the subsequent Transformer fusion layers even when modalities have different numbers of tokens or patches.

We treat modality encoders as fixed feature extractors.
We select strong, publicly available pretrained Transformer encoders that provide semantically meaningful token-/patch-level representations, and we keep them frozen to isolate the effect of the interaction-aware fusion layer.
Unless noted, we use the final hidden-layer outputs of each pretrained encoder (before any task-specific classification head) as our feature sequences.
 FL-I$^2$MoE fusion is trained on top of these frozen embeddings with the objective $\mathcal{L}$.
The dashed region in Figure~\ref{fig:overview}(c) summarizes the inference-time explanation pipeline described in Section~\ref{subsec:explain_pipeline}.

\subsection{Inference-time Explanation Pipeline}
\label{subsec:explain_pipeline}
\subsubsection{Expert-wise Feature Attribution}
To identify which feature units each interaction expert relies on, we compute expert-wise importance scores using gradient$\times$ attention rollout (Grad$\times$AttnRoll)~\cite{chefer2021generic}. We select Grad$\times$AttnRoll as our primary attribution signal because it yields the most faithful masking behavior in our experiments (Section~\ref{subsec:faithfulness}). In addition, the same importance scores are used to define the candidate feature-pair set for the Monte Carlo interaction analysis (Section~\ref{subsec:mc_interactions}). We select a high-importance subset of features within each modality and form cross-modal feature-pair candidates from this subset.

We treat FL-I$^2$MoE as a single differentiable network and define a scalar target score $S(x)$ from the final logit of the ground-truth class. For a given expert $F_i$, we aggregate attention maps from its fusion Transformer layers, apply residual-aware rollout across layers, and modulate the resulting rollout with gradients to obtain a token-wise importance vector $\alpha^{(i)} \in \mathbb{R}^{T}$, where $T=\sum_{m=1}^{M}T_m$ is the length of the concatenated feature sequence. We then split $\alpha^{(i)}$ back into modality-specific segments using encoder masks to obtain modality- and expert-wise importance maps. CLS tokens are not selected as candidate features and padding positions are ignored. Finally, we min–max normalize scores over valid positions.

\subsubsection{Monte Carlo Estimation of Feature-level Interactions}
\label{subsec:mc_interactions}
We quantify cross-modal feature interactions on top of the trained FL-I$^2$MoE using Monte Carlo estimates of the Shapley Interaction Index (SII) for synergy and a redundancy-oriented score for substitutable information. Crucially, we focus on the synergy and redundancy experts: their attribution maps define a candidate feature set, making pairwise interaction estimation tractable and targeted.

For a fixed sample $x$ and expert $F_i$, we first construct a feature universe $F$ by taking the top-$\rho$ fraction of features in each modality according to expert-wise Grad$\times$AttnRoll scores. We then restrict our attention to cross-modal pairs $(u,v)$ with $u \neq v$ and estimate their interactions by masking features in $F$.

\paragraph{Synergy via the Shapley Interaction Index.}
The Shapley Interaction Index (SII) extends Shapley values from
individual features to feature pairs~\cite{grabisch1999axiomatic}. Let $F$ be the set of
features and $f : \mathbb{R}^{|F|} \rightarrow \mathbb{R}$ a scalar-valued
model. For a pair $i,j \in F$ with $i \neq j$ and a subset
$S \subseteq F \setminus \{i,j\}$, the second-order interaction term is
\begin{equation}
    \Delta_{ij} f(S)
    \;=\;
    f(S \cup \{i,j\})
    - f(S \cup \{i\})
    - f(S \cup \{j\})
    + f(S).
    \label{eq:delta_ij}
\end{equation}
The SII for $(i,j)$ averages these terms over all coalitions $S$ with
Shapley-style weights:
\begin{equation}
    \mathrm{SII}(i,j)
    \;=\;
    \sum_{S \subseteq F \setminus \{i,j\}}
    \frac{|S|!\,\big(|F| - |S| - 2\big)!}{2\,|F|!}
    \;\Delta_{ij} f(S).
    \label{eq:sii_def}
\end{equation}
Positive values $\mathrm{SII}(i,j) > 0$ indicate that the joint effect of
$i$ and $j$ exceeds the sum of their individual effects (synergy), while
negative values correspond to suppressive or redundant behavior.

In our setting, $f$ is the scalar output $f_i$ of expert $F_i$ for the
true class, and we apply SII to cross-modal pairs $(u,v)$ with $u \neq v$.
Exact computation is infeasible for realistic $|F|$, so we follow
\cite{wang2025multishap} and use a Monte Carlo approximation: for each pair
$(u,v)$, we sample coalitions $S \subseteq F \setminus \{u,v\}$ and
evaluate $f_i$ under four masked configurations
$S$, $S \cup \{u\}$, $S \cup \{v\}$, and $S \cup \{u,v\}$.
Our masking is intended to mirror our faithfulness analysis: all features are zeroed by
default, only features in the active subset are restored, and CLS tokens
are always kept active. We estimate $\mathrm{SII}_i(u,v)$ as the empirical
average of $\Delta_{uv} f_i(S)$ over sampled $S$, and treat pairs with
$\mathrm{SII}_i(u,v) > 0$ as synergistic. 

\paragraph{Redundancy via a redundancy gap.}
Non-positive SII does not distinguish redundancy from general antagonistic effects. Since I$^2$MoE explicitly targets PID-style redundant information, we score redundancy as substitutability: either feature alone should nearly suffice, and the joint configuration should not substantially exceed (synergy) or contradict (antagonism) the stronger single-feature effect.

For a pair $(u,v)$ and coalition $S \subseteq F \setminus \{u,v\}$, define gains relative to the same masked baseline:
\begin{equation}
\small
\begin{aligned}
g_u(S) &= f_i(S \cup \{u\}) - f_i(S),\quad
g_v(S) = f_i(S \cup \{v\}) - f_i(S),\\
g_{uv}(S) &= f_i(S \cup \{u,v\}) - f_i(S),\quad
g_{\max}(S)=\max\{g_u(S), g_v(S)\}.
\end{aligned}
\end{equation}
We then extract a shared positive support term and a deviation term:
\begin{equation}
\small
\mathrm{base}(S)=\max\!\big(0,\min(g_{uv}(S), g_{\max}(S))\big),\quad
\mathrm{span}(S)=\big|g_{uv}(S)-g_{\max}(S)\big|.
\end{equation}
Averaging over Monte Carlo samples of $S$ yields $\mathrm{base\_mean}=\mathbb{E}_S[\mathrm{base}(S)]$ and $\mathrm{span\_mean}=\mathbb{E}_S[\mathrm{span}(S)]$, and we define:
\begin{equation}
R_{\mathrm{red},i}(u,v)=\frac{\mathrm{base\_mean}(u,v)}{1+\mathrm{span\_mean}(u,v)}.
\end{equation}
This score is large when (i) at least one of $u$ or $v$ provides strong positive evidence (large base), and (ii) the joint effect is close to the best single-feature effect (small span), matching PID-style redundancy as shared, substitutable evidence.

\section{Experiments}
\subsection{Experimental Setup}
\subsubsection{Datasets and tasks}
We evaluate on three multimodal classification benchmarks:
MM-IMDb (movie poster image + plot text, multi-label genre prediction)~\cite{liang2021multibench},
ENRICO (mobile app GUI screenshot + wireframe, multi-class screen type prediction)~\cite{liang2021multibench},
and MMHS150K (tweet text + image, binary hate-speech detection)~\cite{gomez2020exploring}.
For MM-IMDb and ENRICO, we follow the preprocessing and task definitions used in the original I$^2$MoE~\cite{xin2025i2moe} implementation. For MMHS150K, we adopt the preprocessing and train/validation/test protocol of the MMHS150K paper~\cite{gomez2020exploring}.

\subsubsection{Model variants}
Across these datasets, we evaluate the three architectures introduced in Section~\ref{sec:method}:
(i) a dense Transformer with modality-specific encoders followed by a shared fusion Transformer, 
(ii) the original I$^2$MoE~\cite{xin2025i2moe}, and
(iii) our proposed FL-I$^2$MoE with fine-grained token or patch encoders.

\subsubsection{Training and evaluation}
For MM-IMDb and ENRICO, we reuse the original I$^2$MoE training setup~\cite{xin2025i2moe}. The only change is the temperature parameter (temperature\_rw) on MM-IMDb, which we slightly adjust on a validation set to account for the different feature encoders. For MMHS150K, we adopt a comparable Transformer / I$^2$MoE configuration adapted to the binary setting. Each configuration is trained with three random seeds, and we report mean $\pm$ standard deviation over the runs. For MM-IMDb we measure both micro- and macro-F$_1$; for ENRICO we report accuracy; and for MMHS150K we report accuracy and macro-F$_1$. For computationally expensive interaction analyses
(Monte Carlo SII and redundancy gap; Fig.~\ref{fig:mc_alignment_all} and Table~\ref{tab:pair_main}),
we use one FL-I$^2$MoE checkpoint per dataset.

\subsubsection{Backbone and baseline models}
For all three datasets, our dense Transformer and FL-I$^2$MoE share the same multimodal backbones, and differ only in the fusion/interaction module. We treat all modality encoders in these two models as frozen feature extractors; only the fusion/interaction modules are trained.

On MM-IMDb, movie posters are encoded with the CLIP vision backbone ViT-B/16 (``clip-vit-base-patch16'')~\cite{radford2021learning,dosovitskiy2020image}, and plot summaries are encoded with RoBERTa-base~\cite{liu2019roberta}. On ENRICO, both UI screenshots and wireframe renderings are encoded with the same CLIP ViT-B/16 image encoder (``clip-vit-base-patch16'') and treated as two visual modalities. For MMHS150K, we follow the CLIP setting and use the CLIP ViT-B/32 image encoder together with its paired text encoder (``clip-vit-base-patch32'') for tweet images and text. The original I$^2$MoE instead follows the modality encoders used in~\cite{xin2025i2moe}, where each encoder outputs a single pooled vector per modality before the interaction module. A comparison between the original I$^2$MoE and FL-I$^2$MoE is reported in Sec.~\ref{sec:ablation}.

\subsection{Faithfulness of Expert-wise Importance}
\label{subsec:faithfulness}
We assess the faithfulness of expert-wise attributions using a top-$K\%$ masking protocol~\cite{petsiuk2018rise,fong2017interpretable}, applied to both the dense Transformer and the FL-I$^2$MoE. We evaluate three standard attribution methods: attention rollout (AttnRoll)~\cite{abnar2020quantifying}, gradient$\times$attention rollout (Grad$\times$AttnRoll)~\cite{chefer2021generic}, and integrated gradients (IG)~\cite{sundararajan2017axiomatic}. For each test sample, attribution method, and modality, we obtain a feature-level importance map, restrict to valid (non-padding) positions, and select the top $K\%$ tokens according to their importance scores, for $K \in \{5,10,15,20,25,30\}$. CLS tokens are excluded from the candidate set. Masking is implemented by zeroing only the selected feature representations while keeping sequence length and attention masks unchanged. To validate whether masking truly targets important features rather than just any features, we also evaluate a random masking baseline. For this baseline, we mask the same number of valid features but choose them uniformly at random. For each configuration, we evaluate the primary task metric $M$ on the unmasked test set ($M^{\text{base}}$) and under masking ($M^{\text{masked}}(K)$), and summarize faithfulness via the drop curve
\[
\Delta M(K) = M^{\text{base}} - M^{\text{masked}}(K),
\]
averaged over $K \in \{5,10,15,20,25,30\}$ and three random seeds to yield a single summary value per dataset and attribution method.

\begin{figure}[t]
    \centering
    % --- MM-IMDb ---
    \begin{subfigure}[b]{0.32\textwidth}
        \centering
        \begin{tikzpicture}
        \begin{axis}[
            ybar,
            ymin=0, ymax=35,
            width=\linewidth,
            height=4.0cm,
            bar width=5pt,
            enlarge x limits=0.2,
            symbolic x coords={Random,AttnRoll,IG,Grad$\times$AttnRoll},
            xtick=data,
            xticklabel style={font=\scriptsize, rotate=45, anchor=east},
            ylabel={Micro-F$_1$ drop},
            ylabel style={font=\scriptsize},
            tick label style={font=\scriptsize},
            legend to name=dropLegend,
            legend style={legend columns=-1,
                          /tikz/every even column/.append style={column sep=0.5em}},
            legend image code/.code={
      \draw[#1] (0cm,-0.08cm) rectangle (0.28cm,0.08cm);
    }
        ]
            % MM-IMDb: dense Transformer, I^2MoE
            \addplot coordinates {
                (Random,1.6)
                (AttnRoll,11.3)
                (IG,10.3)
                (Grad$\times$AttnRoll,19.5)
            };

            \addplot coordinates {
                (Random,18.2)
                (AttnRoll,32.3)
                (IG,31.6)
                (Grad$\times$AttnRoll,28.4)
            };
            \legend{dense Transformer, FL-I$^2$MoE}
            
        \end{axis}
        \end{tikzpicture}
        \caption{MM-IMDb}
    \end{subfigure}
    \hfill
    % --- ENRICO ---
    \begin{subfigure}[b]{0.32\textwidth}
        \centering
        \begin{tikzpicture}
        \begin{axis}[
            ybar,
            ymin=0, ymax=32,
            width=\linewidth,
            height=4.0cm,
            bar width=5pt,
            enlarge x limits=0.2,
            symbolic x coords={Random,AttnRoll,IG,Grad$\times$AttnRoll},
            xtick=data,
            xticklabel style={font=\scriptsize, rotate=45, anchor=east},
            ylabel={Acc. drop},
            ylabel style={font=\scriptsize},
            tick label style={font=\scriptsize},
        ]
            % ENRICO: dense Transformer, I^2MoE
            \addplot coordinates {
                (Random,0.6)
                (AttnRoll,2.9)
                (IG,2.3)
                (Grad$\times$AttnRoll,7.3)
            };
            \addplot coordinates {
                (Random,7.1)
                (AttnRoll,17.4)
                (IG,20.3)
                (Grad$\times$AttnRoll,30.6)
            };
        \end{axis}
        \end{tikzpicture}
        \caption{ENRICO}
    \end{subfigure}
    \hfill
    % --- MMHS150K ---
    \begin{subfigure}[b]{0.32\textwidth}
        \centering
        \begin{tikzpicture}
        \begin{axis}[
            ybar,
            ymin=0, ymax=10,
            width=\linewidth,
            height=4.0cm,
            bar width=5pt,
            enlarge x limits=0.2,
            symbolic x coords={Random,AttnRoll,IG,Grad$\times$AttnRoll},
            xtick=data,
            xticklabel style={font=\scriptsize, rotate=45, anchor=east},
            ylabel={Acc. drop},
            ylabel style={font=\scriptsize},
            tick label style={font=\scriptsize},
        ]
            % MMHS150K: dense Transformer, I^2MoE
            \addplot coordinates {
                (Random,0.1)
                (AttnRoll,2.2)
                (IG,2.2)
                (Grad$\times$AttnRoll,3.9)
            };
            \addplot coordinates {
                (Random,0.1)
                (AttnRoll,4.3)
                (IG,3.5)
                (Grad$\times$AttnRoll,7.6)
            };
        \end{axis}
        \end{tikzpicture}
        \caption{MMHS150K}
    \end{subfigure}

    \vspace{0.3em}
    \pgfplotslegendfromname{dropLegend}

    \caption{Average performance drop when masking the top-$K\%$ features
    according to different attribution methods (Random, AttnRoll, IG, and Grad$\times$AttnRoll).}
    \label{fig:avg_drop_overall_pgf}
\end{figure}
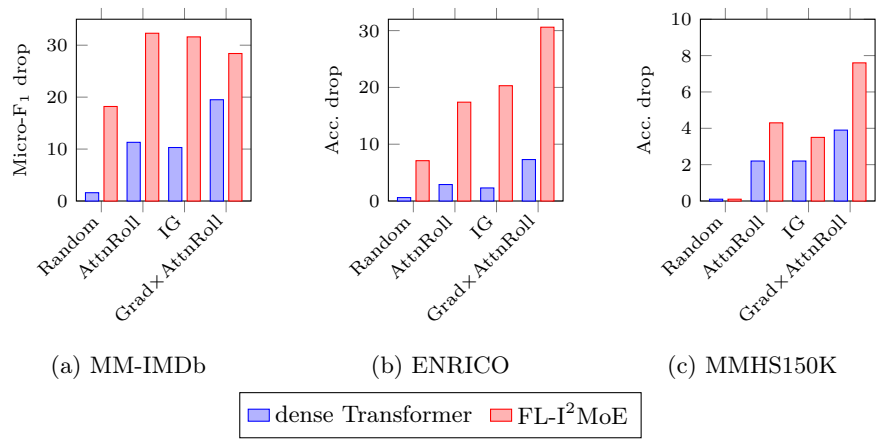

Across all datasets, masking features according to attribution scores leads to larger performance drops than masking the same number of randomly selected features, for both the dense Transformer and the FL-I$^2$MoE in Figure~\ref{fig:avg_drop_overall_pgf}. This confirms that, in both architectures, the attribution maps identify features that are causally important for the prediction rather than arbitrarily selected features.

Moreover, across all datasets, the FL-I$^2$MoE consistently suffers larger drops from important feature masking than the dense Transformer in Figure~\ref{fig:avg_drop_overall_pgf}. In masking evaluations, this pattern is interpreted as stronger faithfulness. A relatively small set of high-scoring features already suffices to substantially disrupt the prediction, which suggests that the interaction-specific experts concentrate importance on a focused subset of important features instead of spreading 
it diffusely across many features.

\subsection{Alignment with Monte Carlo Feature Interactions}
We next relate expert-wise importance to the Monte Carlo identification of interaction metrics from Section~\ref{subsec:mc_interactions}. For each dataset and expert $F_i$, we first construct a candidate feature universe by taking the top-$30\%$ features per modality under Grad$\times$AttnRoll. Within this universe, we partition features into three disjoint importance bins: top-10\%, 11–20\%, and 21–30\%. For each bin, we form cross-modal pairs $(u,v)$ by combining features from the same bin across modalities, which yields three sets of cross-modal pairs built from increasingly less important features. For every pair in each bin, we estimate $\mathrm{SII}_i(u,v)$ (for the synergy expert) or $R_{\mathrm{red},i}(u,v)$ (for the redundancy expert) using the same Monte Carlo masking procedure as in Section~\ref{subsec:mc_interactions}, with four randomly sampled context subsets $S$ per pair to keep the computational cost manageable. We then summarize each bin by reporting the mean interaction value over the top-$q\%$ pairs under the corresponding metric ($q \in \{5,10,20\}$).

\begin{figure}[t]
\centering

\resizebox{0.85\linewidth}{!}{%
\begin{minipage}{\linewidth}

% ========= Top row: Synergy (mean SII) =========

% --- MM-IMDb (Synergy) ---
\begin{subfigure}[b]{0.31\textwidth}
\centering
\begin{tikzpicture}
\begin{axis}[
    ybar,
    ymin=0, ymax=4.0,
    width=\linewidth,
    height=3.5cm,
    bar width=4pt,
    enlarge x limits=0.25,
    symbolic x coords={top-10,11--20,21--30},
    xtick=data,
    xticklabel style={font=\scriptsize, rotate=30, anchor=east},
    ylabel={mean SII ($\times 10^{-3}$)},
    ylabel style={font=\scriptsize},
    tick label style={font=\scriptsize},
    title={MM-IMDb},
    legend to name=mcLegend,
    legend style={
        font=\scriptsize,
        legend columns=3,
        /tikz/every even column/.append style={column sep=0.5em}
    },
    legend image code/.code={
        \draw[#1] (0cm,-0.08cm) rectangle (0.28cm,0.08cm);
    }
]
    \addplot coordinates {
        (top-10,2.64)
        (11--20,1.23)
        (21--30,1.28)
    };
    \addplot coordinates {
        (top-10,1.50)
        (11--20,0.763)
        (21--30,0.787)
    };
    \addplot coordinates {
        (top-10,0.855)
        (11--20,0.466)
        (21--30,0.475)
    };
    \legend{top-5\%, top-10\%, top-20\%}
\end{axis}
\end{tikzpicture}
\end{subfigure}\hspace{0.01\textwidth}%
%
% --- ENRICO (Synergy) ---
\begin{subfigure}[b]{0.31\textwidth}
\centering
\begin{tikzpicture}
\begin{axis}[
    ybar,
    ymin=0, ymax=4.0,
    width=\linewidth,
    height=3.5cm,
    bar width=4pt,
    enlarge x limits=0.25,
    symbolic x coords={top-10,11--20,21--30},
    xtick=data,
    xticklabel style={font=\scriptsize, rotate=30, anchor=east},
    ylabel={mean SII ($\times 10^{-3}$)},
    ylabel style={font=\scriptsize},
    tick label style={font=\scriptsize},
    title={ENRICO}
]
    \addplot coordinates {
        (top-10,3.54)
        (11--20,2.02)
        (21--30,1.31)
    };
    \addplot coordinates {
        (top-10,2.46)
        (11--20,1.29)
        (21--30,0.83)
    };
    \addplot coordinates {
        (top-10,1.62)
        (11--20,0.813)
        (21--30,0.53)
    };
\end{axis}
\end{tikzpicture}
\end{subfigure}\hspace{0.01\textwidth}%
%
% --- MMHS150K (Synergy) ---
\begin{subfigure}[b]{0.31\textwidth}
\centering
\begin{tikzpicture}
\begin{axis}[
    ybar,
    ymin=0, ymax=4.0,
    width=\linewidth,
    height=3.5cm,
    bar width=4pt,
    enlarge x limits=0.25,
    symbolic x coords={top-10,11--20,21--30},
    xtick=data,
    xticklabel style={font=\scriptsize, rotate=30, anchor=east},
    ylabel={mean SII ($\times 10^{-3}$)},
    ylabel style={font=\scriptsize},
    tick label style={font=\scriptsize},
    title={MMHS150K}
]
    \addplot coordinates {
        (top-10,3.82)
        (11--20,1.78)
        (21--30,1.44)
    };
    \addplot coordinates {
        (top-10,3.18)
        (11--20,1.31)
        (21--30,1.02)
    };
    \addplot coordinates {
        (top-10,2.46)
        (11--20,0.882)
        (21--30,0.648)
    };
\end{axis}
\end{tikzpicture}
\end{subfigure}

\vspace{0.2em}

% ========= Bottom row: Redundancy (mean R_red) =========

% --- MM-IMDb (Redundancy) ---
\begin{subfigure}[b]{0.31\textwidth}
\centering
\begin{tikzpicture}
\begin{axis}[
    ybar,
    ymin=0, ymax=1.2,
    width=\linewidth,
    height=3.5cm,
    bar width=4pt,
    enlarge x limits=0.25,
    symbolic x coords={top-10,11--20,21--30},
    xtick=data,
    xticklabel style={font=\scriptsize, rotate=30, anchor=east},
    ylabel={mean $R_{\text{red}}$},
    ylabel style={font=\scriptsize},
    tick label style={font=\scriptsize},
    title={MM-IMDb}
]
    \addplot coordinates {
        (top-10,1.13)
        (11--20,0.28)
        (21--30,0.19)
    };
    \addplot coordinates {
        (top-10,0.72)
        (11--20,0.19)
        (21--30,0.13)
    };
    \addplot coordinates {
        (top-10,0.44)
        (11--20,0.12)
        (21--30,0.08)
    };
\end{axis}
\end{tikzpicture}
\end{subfigure}\hspace{0.01\textwidth}%
%
% --- ENRICO (Redundancy) ---
\begin{subfigure}[b]{0.31\textwidth}
\centering
\begin{tikzpicture}
\begin{axis}[
    ybar,
    ymin=0, ymax=0.45,
    width=\linewidth,
    height=3.5cm,
    bar width=4pt,
    enlarge x limits=0.25,
    symbolic x coords={top-10,11--20,21--30},
    xtick=data,
    xticklabel style={font=\scriptsize, rotate=30, anchor=east},
    ylabel={mean $R_{\text{red}}$},
    ylabel style={font=\scriptsize},
    tick label style={font=\scriptsize},
    title={ENRICO}
]
    \addplot coordinates {
        (top-10,0.40)
        (11--20,0.14)
        (21--30,0.09)
    };
    \addplot coordinates {
        (top-10,0.31)
        (11--20,0.11)
        (21--30,0.08)
    };
    \addplot coordinates {
        (top-10,0.23)
        (11--20,0.09)
        (21--30,0.06)
    };
\end{axis}
\end{tikzpicture}
\end{subfigure}\hspace{0.01\textwidth}%
%
% --- MMHS150K (Redundancy) ---
\begin{subfigure}[b]{0.31\textwidth}
\centering
\begin{tikzpicture}
\begin{axis}[
    ybar,
    ymin=0, ymax=0.25,
    width=\linewidth,
    height=3.5cm,
    bar width=4pt,
    enlarge x limits=0.25,
    symbolic x coords={top-10,11--20,21--30},
    xtick=data,
    xticklabel style={font=\scriptsize, rotate=30, anchor=east},
    ylabel={mean $R_{\text{red}}$},
    ylabel style={font=\scriptsize},
    tick label style={font=\scriptsize},
    title={MMHS150K}
]
    \addplot coordinates {
        (top-10,0.21)
        (11--20,0.04)
        (21--30,0.03)
    };
    \addplot coordinates {
        (top-10,0.16)
        (11--20,0.03)
        (21--30,0.02)
    };
    \addplot coordinates {
        (top-10,0.10)
        (11--20,0.02)
        (21--30,0.01)
    };
\end{axis}
\end{tikzpicture}
\end{subfigure}

\vspace{0.2em}

% ========= Shared legend =========
\begin{center}
\pgfplotslegendfromname{mcLegend}
\end{center}
\end{minipage}
}

\caption{Alignment between expert-wise importance and Monte Carlo interaction metrics.
Top: mean SII (synergy expert). Bottom: mean redundancy gap
$R_{\text{red}}$ (redundancy expert). Bars compare
importance bins (x-axis) and top-$q\%$ sets.}
\label{fig:mc_alignment_all}

\end{figure}
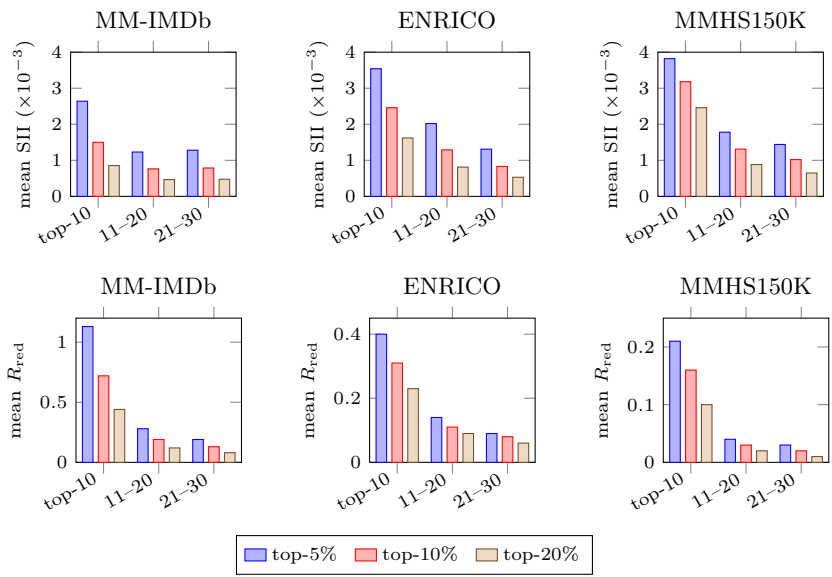

Figure~\ref{fig:mc_alignment_all} shows that, across all datasets, attribute bins built from more important features consistently achieve larger positive SII for the synergy expert and larger redundancy gaps for the redundancy expert. Although the absolute magnitudes vary by dataset, the ordering of bins is stable: feature pairs formed from higher-importance features also exhibit stronger measured interactions. Notably, although the Monte Carlo interaction estimates are computed from a single checkpoint per dataset, this importance-bin ordering is observed consistently on three datasets. We view this as evidence that the proposed interaction metrics behave coherently: rather than assigning high values to arbitrary cross-modal pairs, the Monte Carlo SII and redundancy-gap scores increase in a systematic way as expert-wise feature importance increases.

\subsection{Pair-level Masking of Synergistic and Redundant Interactions}
\label{sec:pair_deletion}

We test whether high-scoring feature pairs are causally important beyond single-feature importance. For each expert $F_i$ and each dataset, we restrict candidates to the top-10\% features, enumerate cross-modal pairs, and mask $K{=}5\%$ of these pairs. For the synergy expert, we compare two selection rules: (i) masking randomly sampled cross-modal pairs from the candidate pool, and (ii) masking pairs ranked by $\mathrm{SII}_i(u,v)$. For the redundancy expert, we use the same setup but rank pairs by $R_{\mathrm{red},i}(u,v)$ instead. In all cases, masking a pair means jointly zeroing the corresponding two features while leaving all other features untouched. Table~\ref{tab:pair_main} summarizes the results. Across all three datasets, masking SII-ranked synergy pairs or redundancy-gap–ranked pairs consistently causes larger performance drops than masking randomly chosen pairs under the same pair budget.

Because all pairs are drawn from the same top-10\% feature pool, the gap between random and interaction-ranked masking reflects pair-level interaction rather than marginal importance alone. Thus, SII and redundancy-gap scores identify cross-modal pairs whose joint or substitutable effects are especially important for prediction.

\begin{table}[t]
\centering
\small
\setlength{\tabcolsep}{4pt}
\caption{
Pair masking at $K{=}5\%$ using candidate pairs formed from the top-$10\%$ features per modality. For each dataset, we report performance after masking random pairs versus pairs selected
by SII (synergy) or by redundancy-gap scores.
}
\label{tab:pair_main}
\vspace{0.1\baselineskip}
\resizebox{0.85\linewidth}{!}{
\begin{tabular}{lcccccc}

\toprule
& \multicolumn{2}{c}{MM-IMDb (Micro F1)}
& \multicolumn{2}{c}{ENRICO (Acc.)}
& \multicolumn{2}{c}{MMHS150K (Acc.)} \\
\cmidrule(lr){2-3}\cmidrule(lr){4-5}\cmidrule(lr){6-7}
Masking setting
& Syngy & Rdncy
& Syngy & Rdncy
& Syngy & Rdncy \\
\midrule
Original     & 0.6789 & 0.6789 & 0.5479 & 0.5479 & 0.7004 & 0.7004 \\
Random pairs & 0.5703 & 0.6136 & 0.5137 & 0.5240 & 0.6796 & 0.6839 \\
\rowcolor{gray!15}
Interaction pairs
             & 0.5600 & 0.5932 & 0.4795 & 0.5171 & 0.6708 & 0.6745 \\
\bottomrule
\end{tabular}
}
\end{table}

\subsection{Qualitative Case Study}
\label{subsec:qual_mmimdb}

We now revisit the MM-IMDb example (\emph{Four Minutes}, true labels: \emph{drama}, \emph{music}) introduced in Figure~\ref{fig:mmimdb_qual_input}, and qualitatively compare explanations from a dense Transformer and the FL-I$^2$MoE. Both models partially match the labels; our goal here is to contrast the structure of their explanations rather than to optimize predictive performance on this single instance.

\begin{figure*}[t]
    \centering
    \includegraphics[width=\textwidth]{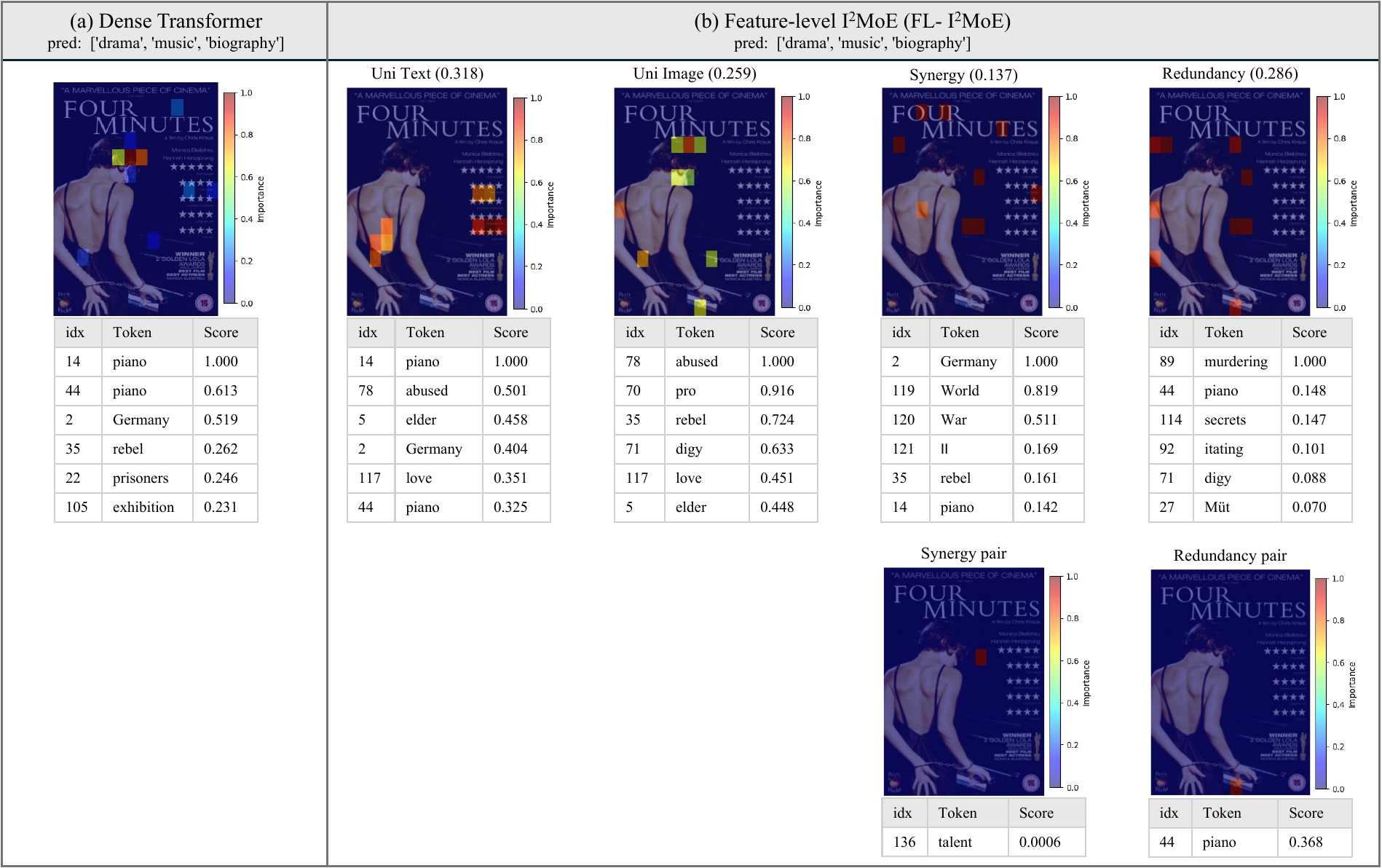}
    \caption{
    Qualitative comparison on an MM-IMDb example.
    (a) dense Transformer with separate image and text
    saliency maps. 
    (b) FL-I$^2$MoE. Top row: uni-text, uni-image,
    synergy, and redundancy experts. Bottom row: token-patch pairs by Monte Carlo interaction metrics for high synergy (SII) or redundancy
    gap.
    }
    \label{fig:qual_mmimdb}
\end{figure*}

Figure~\ref{fig:qual_mmimdb}(a) shows separate image and text Grad$\times$AttnRoll maps for the dense Transformer: text saliency highlights music-related and emotionally charged tokens, while image saliency mostly focuses on the face and upper body, offering limited direct evidence for music. Figure~\ref{fig:qual_mmimdb}(b) shows FL-I$^2$MoE. Uni-modal experts behave similarly to the dense Transformer but allocate more mass to piano regions, while synergy and redundancy experts highlight distinct cross-modal token-patch pairs with large SII or redundancy-gap scores. This makes it possible to inspect which specific patches and tokens interact and how they contribute (as unique, synergistic, or redundant evidence), rather than only seeing a single undifferentiated saliency map per modality. In this way, the interaction-aware architecture turns modality-wise saliency into 
structured explanations that separate unique, synergistic, and redundant evidence for the same prediction.

\section{Ablation Study}
\subsection{Effect of Feature-level Encoders}
\label{sec:ablation}

We compare the original I$^2$MoE with the proposed feature-level variant under identical multimodal backbones. Table~\ref{tab:encoder_comparison} reports the classification results. On MM-IMDb, the FL-I$^2$MoE improves over the pooled-encoder variant by about $+7.5$ micro-F$_1$ and $+5.1$ macro-F$_1$. On ENRICO, accuracy increases from $45.43 \pm 1.13$ to $51.26 \pm 2.17$, a gain of roughly $+5.8$ points. 

These ablations show that replacing pooled encoders with fine-grained encoders inside I$^2$MoE does not degrade task performance and can yield modest improvements, while primarily serving as a structured interpretability layer that can be plugged into multimodal Transformers without sacrificing accuracy.

\begin{table}[t]
\centering
\small
\setlength{\tabcolsep}{4pt}
\caption{Comparison between the original I$^2$MoE and the proposed FL-I$^2$MoE.
Results are reported as mean $\pm$ standard deviation over three runs.}
\label{tab:encoder_comparison}
\vspace{0.1\baselineskip}
\resizebox{0.85\linewidth}{!}{
\begin{tabular}{lccccc}
\toprule
& \multicolumn{2}{c}{MM-IMDb} & \multicolumn{2}{c}{ENRICO} \\
\cmidrule(lr){2-3} \cmidrule(lr){4-5}
Metric & Original I$^2$MoE & FL-I$^2$MoE & Original I$^2$MoE & FL-I$^2$MoE \\
\midrule
Micro-F1 & $60.33 \pm 0.47$ & $67.85 \pm 0.45$ & -- & -- \\
Macro-F1 & $51.21 \pm 0.57$ & $56.35 \pm 0.15$ & -- & -- \\
Accuracy & -- & -- & $45.43 \pm 1.13$ & $51.26 \pm 2.17$ \\
\bottomrule
\end{tabular}
}
\end{table}

\section{Conclusion}
We present a feature-pair–level framework for explaining cross-modal interactions beyond modality-wise saliency, enabling identification of complementary or substitutable evidence. Our FL-I$^2$MoE operates on token/patch sequences from frozen pretrained encoders, allowing fine-grained separation of unique, synergistic, and redundant information. Experiments on MM-IMDb, ENRICO, and MMHS150K show that attribution-based masking leads to larger performance drops than random masking, supporting faithfulness, and that FL-I$^2$MoE produces more concentrated importance than a dense Transformer, indicating more concentrated and faithful explanations. We further introduce Monte Carlo interaction metrics (SII and redundancy gap) and show that high-scoring feature pairs are causally important through pair-level masking. Future work will focus on improving Monte Carlo efficiency and extending to more modalities and higher-order interactions.

\section*{Acknowledgements}
This work was supported by the Natural Sciences and Engineering Research Council of Canada (NSERC) Collaborative Research and Training Experience (CREATE) program “From Data to Decision” (FD2D). This research was also supported by the Alberta Machine Intelligence Institute (Amii), NSERC (including grants DGECR-2022-00369 and RGPIN-2022-0346), and Alberta Innovates (Enabling Better Health through Artificial Intelligence (AI-Better Health) Program)

\bibliographystyle{splncs04}
\bibliography{references}

\end{document}